\documentclass[letterpaper, 10 pt, conference]{ieeeconf}  

\IEEEoverridecommandlockouts                              

\usepackage{graphics} 
\usepackage{amsmath} 
\usepackage{amssymb}  

\usepackage{graphicx} 
\usepackage{caption}
\usepackage{subcaption}
\usepackage{cite}
\usepackage{balance}
\usepackage{hyperref}

\usepackage{todonotes}
\newcommand{\rev}[1]{\textcolor{black}{#1}}

\title{\LARGE \bf
Radio-based Multi-Robot Odometry and Relative Localization
}

\author{Andrés Martínez-Silva$^{1}$, David Alejo$^{2}$, Luis Merino$^{1}$ and Fernando Caballero$^{1}$ 
\thanks{*This work was supported by the grants PICRAH 4.0 0 (PLEC2023-010353): funded by the Spanish Ministry of Science and Innovation and the Spanish Research Agency (MCIN/AEI/10.13039/501100011033); and INSERTION (PID2021-127648OB-C31), funded by “Agencia Estatal de Investigación – Ministerio de Ciencia, Innovación y Universidades” and the “European Union NextGenerationEU/PRTR”.}%
\thanks{$^{1}$Andrés Martínez-Silva, Fernando Caballero and Luis Merino are with the Service Robotics Laboratory. University Pablo de Olavide, CP 41013, Seville, Spain.\texttt{\{amarsil,fcaballero,lmercab\}@upo.es}}%
\thanks{$^{2}$David Alejo is with the Department of System Engineering and Automation, University of Seville, Spain. \texttt{dalejo@us.es}}
}

\begin{document}

\maketitle
\thispagestyle{empty}
\pagestyle{empty}

\begin{abstract}

Radio-based methods such as Ultra-Wideband (UWB) and RAdio Detection And Ranging (radar), which have traditionally seen limited adoption in robotics, are experiencing a boost in popularity thanks to their robustness to harsh environmental conditions and cluttered environments. This work proposes a multi-robot UGV-UAV localization system that leverages the two technologies with inexpensive and readily-available sensors, such as Inertial Measurement Units (IMUs) and wheel encoders, to estimate the relative position of an aerial robot with respect to a ground robot. The first stage of the system pipeline includes a nonlinear optimization framework to trilaterate the location of the aerial platform based on UWB range data, and a radar pre-processing module with loosely coupled ego-motion estimation which has been adapted for a multi-robot scenario. Then, the pre-processed radar data as well as the relative transformation are fed to a pose-graph optimization framework with odometry and inter-robot constraints. The system, implemented for the Robotic Operating System (ROS 2) with the Ceres optimizer, has been validated in Software-in-the-Loop (SITL) simulations and in a real-world dataset. The proposed relative localization module outperforms state-of-the-art closed-form methods which are less robust to noise. Our SITL environment includes a custom Gazebo plugin for generating realistic UWB measurements modeled after real data.  Conveniently, the proposed factor graph formulation makes the system readily extensible to full Simultaneous Localization And Mapping (SLAM).  Finally, all the code and experimental data is publicly available to support reproducibility and to serve as a common open dataset for benchmarking.

\end{abstract}

\section{Introduction}

Heterogeneous multi-robot systems have gained increasing popularity in the last few years. Deploying a team of different robots that work cooperatively - i.e. an Unmanned Ground Vehicle (UGV) and an Unmanned Aerial Vehicle (UAV) - can bring many benefits: one robot may be allowed to access locations that are blocked to the other, enabling the team to cover more terrain in less time. One key challenge is the localization of the robots in the same frame. This paper addresses how information sharing between robots via range-based radio sensors in GPS-denied environments potentially enables more robust multi-robot localization. 

In a multi-robot team, each robot should be able to determine its own position and the relative position of the other robot. Although most of the research focuses on vision- or LiDAR-based methods, the unstructured layout of some areas, together with unfavorable environmental conditions such as fog, dust, or poor lighting  critically impact the performance of these methods \cite{Brooker2007,Starr2014}. Radio-based localization, in contrast, is gaining traction in the literature, especially Ultra-Wideband (UWB) and RAdio Detection And Ranging (radar) methods. UWB sensors provide high-frequency range measurements between anchors and tags with centimetric precision. They are also lightweight, cost-effective, and robust to harsh environmental conditions. Standard UWB-based localization systems require the anchors to be installed in the environment for effective trilateration, but this is often impractical. Deploying a team of robots, on the other hand, makes it possible to place the transceivers onboard to obtain inter-robot distances that can be fused with odometry measurements to determine the positions of both robots in real time. On the other hand, radars are able to estimate the position (in the form of range, azimuth and elevation) and radial velocity of detected objects by leveraging the Doppler effect. However, radar data brings a particular set of challenges regarding sparsity, lower angular and spatial resolution, and multimodal noise coming from reflections or signal scattering, among others \cite{elena2024loosely}.

While there is extensive research on relative transformation estimation based on range measurements, there is very little work on UWB-based heterogeneous multi-robot localization systems that: 1) leverage multiple on-board transceivers for redundancy and enhanced state observability, 2) fuse UWB with radar odometry to localize both robots with respect to a fixed frame, 3) involve robots operating in 3D space and 4) can run in real time.

The main contribution of this paper is a radio-based multi-robot UGV-UAV localization system that comprises three fundamental modules:

\begin{itemize}
\item A Nonlinear Least Squares (NLS) optimization framework to compute the relative transformation between the odometry frames of a UAV and a UGV based on multiple anchor-tag distance measurements, in real time.
\item A pose-graph optimization framework that simultaneously optimizes the poses of both platforms by fusing all available sensor data, including the inter-robot transformation, the pre-processed radar point-clouds and other readily-available sensors such as IMUs or wheel encoders, in real time.
\item A simulator plugin for Gazebo Harmonic that replicates UWB range measurements based on a measurement model that has been validated against real sensor data. This plugin is compatible with multi-robot setups in the open-source PX4 Software-In-The-Loop (SITL) framework, narrowing the gap between simulation and real experimentation. 
\end{itemize}

The system was validated in SITL simulations, and in a real-world experiment involving a UAV and a UGV. All the code and the dataset are publicly available 
\footnote{\href{https://github.com/robotics-upo/mr-radio-localization}{https://github.com/robotics-upo/mr-radio-localization}}

The remainder of the paper is structured as follows: Section \ref{sec:related} reviews previous work on radio-based localization in multi-robot systems. Section \ref{sec:formulation} formalizes our approach by describing the proposed localization system, focusing on our three main contributions. Section \ref{sec:experiments} validates our implementation in simulation and a real dataset. Finally, Section \ref{sec:conclusions} presents some final remarks.

\section{Related work}\label{sec:related}

The problem of estimating inter-robot relative transformations based on distance measurements has been extensively studied in \cite{trawny2010interrobot, trawny20093d}, where the authors propose an algebraic method to compute a 6 Degrees of Freedom (DOF) rigid transformation between robot frames. A closed-form solution can be found from ten distance measurements, and a discrete set of 40 solutions from as few as six. In \cite{molina2019unique}, the authors particularize the case to UWB sensors and achieve a 4-DOF closed-form solution from just six measurements. However, these methods are highly sensitive to outliers and noise and typically require further refinement with Nonlinear Least Squares (NLS). More recently, the authors of \cite{nguyen2023relative} perform an exhaustive analysis on 4-DOF relative frame transformation (RFT) estimation using UWB and visual-inertial odometry, proposing a global estimation approach to improve NLS initialization. However, these methods assume only one sensor per platform and do not exploit redundancy and frequency of multiple onboard measurements. Other research has considered up to four onboard UWB nodes \cite{cao2021relative} to compute instantaneous body frame transformations, though it relies on the availability of sufficient simultaneous measurements. In heterogeneous teams of robots, this assumption is often challenged due to limitations in sensor placement, communication quality and noise.

Radar-based localization and SLAM have been primarily explored in the context of autonomous driving, due to its advantages over other sensor modalities in adverse weather conditions. In \cite{cen2018precise}, the authors present a radar-only motion estimation system using a Frequency Modulated Continuous Wave (FMCW) radar without Doppler information. In \cite{holder2019real}, a real-time pose-graph SLAM system that leverages Iterative Closest Point (ICP)-based scan-matching to obtain relative transformations between consecutive radar detections is introduced. To localize the vehicle, the odometry component implements an Unscented Kalman Filter (UKF) that fuses wheel speed, yaw rate, steering wheel angle and velocity of the radar sensor. The latter is obtained from radial (Doppler) velocities using a radar ego-motion estimation algorithm \cite{kellner2013instantaneous}. In robotics, radar-based systems have only recently seen significant development. Notably, the authors of \cite{zhang20234dradarslam} introduced an open-source Robot Operating System (ROS) package that implements real-time 6-DOF SLAM for 4D radars. More recently, the authors of \cite{elena2024loosely} proposed a graph-based radar-inertial odometry system with a loosely coupled ego-motion estimation algorithm for holonomic and non-holonomic ground robots. However, while these methods were designed with individual robot missions in mind, this paper applies the formulation to multi-robot missions by leveraging relative range data coming from UWB transceivers placed onboard two different platforms.  

Finally, cooperative localization involving a team of robots has been widely covered in the past \cite{kim2010multiple, deutsch2016framework, indelman2014multi}. Recently, Brunacci et al. \cite{brunacci} used a model of the Dilution of Precision (DoP) of the UWB sensors to propose an active perception method to increase the accuracy of the relative transform estimation. Shalihan et al. \cite{SHALIHAN2025103410} propose the MR-FLOUR method that uses UWB inter-robot measurements fused with wheel odometry and LiDAR inter-robot detections to enhance the precision of the localization system. However, it is difficult to find relevant work regarding teams of different robots that localize themselves with radio sensors, especially combining UWB and radar in 3-dimensional settings, as most of the methods in the literature focus on teams of UGVs. 

To the best of our knowledge, this is the first work that integrates relative transformation estimation as an inter-robot factor in a multi-robot pose-graph optimization framework, together with radar factors in 3-dimensional settings.

\section{Problem formulation}\label{sec:formulation}

In this Section, we define the challenge to be tackled and introduce the main components of our multi-robot localization system. In Section \ref{sec:relative}, we present our method to estimate the relative transformation between the odometry frames of the UAV and the UGV based on distance measurements. Both robots are equipped with UWB modules, which are placed in fixed locations relative to their body frame of reference. In Section \ref{sec:radar} we introduce the radar odometry module, and finally, in Section \ref{sec:posegraph}, we fuse all the available information in a pose graph used to compute the positions of both robots with respect to a common reference frame.

\subsection{Relative transform estimation}\label{sec:relative}

\begin{figure}[t!]
    \centering
    \includegraphics[width=0.45\textwidth]{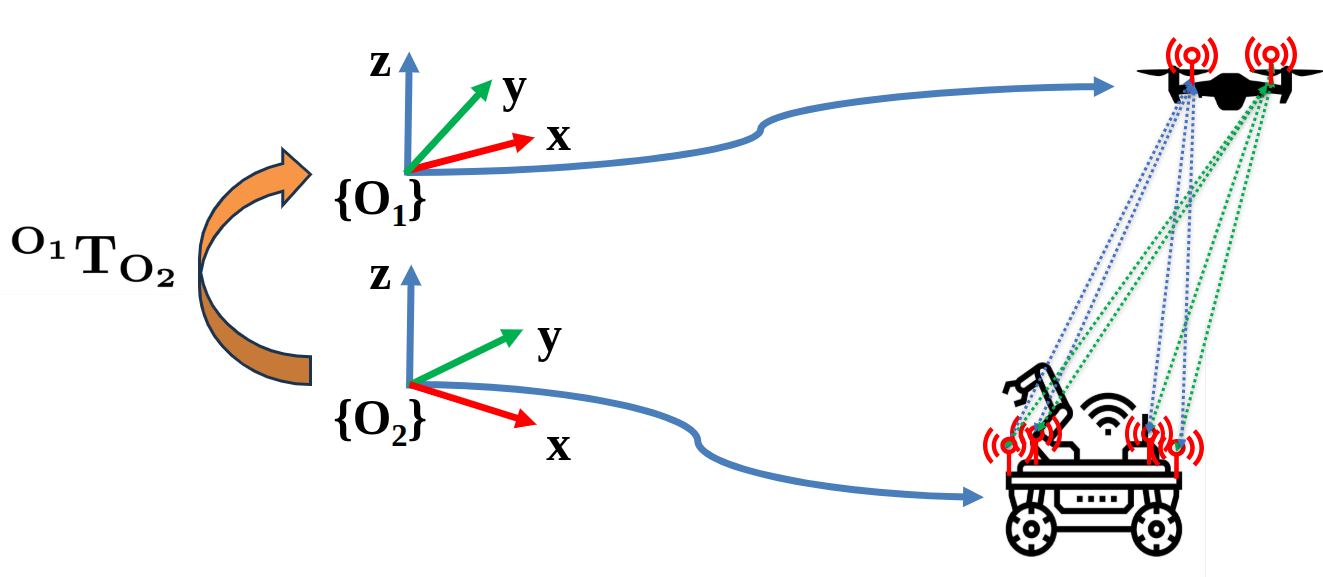}
    \caption{Each robot describes a trajectory in its own local odometry frame while collecting range measurements (dashed lines) between two tags and four anchors. Measurement from different tags are represented in different colors.}
    \label{fig:uwb_formulation}
    \vspace{-5mm}
\end{figure}

Suppose a heterogeneous team of two robots equipped with UWB transceivers moving in proximity, each of them describing a trajectory in their own local reference frame. The UAV (robot 1) is equipped with two tags, and the UGV (robot 2) is equipped with four anchors, \rev{in a configuration that maximizes the distance between the transceivers and clearance} (see Figure \ref{fig:uwb_formulation}).

Let $\mathbf{^{B_{1}}\tilde{p}_{ti}}, i \in\{1,2\}$ be the known homogeneous coordinates of the UWB tag $i$ in the body frame of robot 1, and $\mathbf{^{B_{2}}\tilde{p}_{aj}}, j \in \{1:4\}$ the known homogeneous coordinates of the UWB anchor $j$ in the body frame of robot 2, and let $\mathbf{^{O_{1}}p_{1}}$ and $\mathbf{^{O_{2}}p_{2}}$ be the positions of robots 1 and 2 in their own local odometry frames, respectively. Consequently, the coordinates of tag $i$ in the odometry frame of robot 1, $\mathbf{^{O_{1}}\tilde{p}_{ti}}$, and the coordinates of anchor $j$ in the odometry frame of robot 2, $\mathbf{^{O_{2}}\tilde{p}_{aj}}$, are known.

Let $^{O_{1}}T_{O_{2}}$ be the rigid transformation between the odometry frames of the robots. The predicted distance $\hat{d}$ between tag $i$ and anchor $j$ is defined as follows:

\begin{equation}\label{eq:predicted_range}
\hat{d}(t_{i},a_{j}) = ||\mathbf{^{O_{1}}\tilde{p}_{ti}} - ^{O_{1}}T_{O_{2}}\mathbf{^{O_{2}}\tilde{p}_{aj}}|| + \epsilon\,
\end{equation}

\noindent where $\epsilon \sim \mathcal{N}(0,\,\sigma^{2})$ represents independent, identically distributed Gaussian noise.

The objective is to find the inter-robot transformation $^{O_{1}}T_{O_{2}}$ between the odometry frames of both platforms, which allows us to map the local positions of robot 2 to the local reference frame of robot 1, according to the anchor-tag range measurements $d$. We solve this problem in terms of nonlinear least squares, where the cost function is defined as:

\small
\begin{equation}\label{eq:costfunct}
    J = \min_{^{O_{1}}T_{O_{2}}}\sum_{W}\left(\sum_{i,j}\left(\frac{\hat{d}(t_{i},a_{j}) - d}{\sigma}\right)^{2} + ||\mathbf{e}\left(^{O_{1}}T_{O_{2}}, T_{p}\right)||^{2}_\Sigma\right)
\end{equation}
\normalsize
The first term is a scalar that, for each anchor-tag pair, minimizes the difference between the predicted and the actual measurement, divided by its standard deviation $\sigma$, and the second term anchors the estimated transform to a prior $T_{p}$. 

Finally, $||\mathbf{e}||_\Sigma$ is the squared Mahalanobis distance $\mathbf{e}^{T}\Sigma^{-1}\mathbf{e}$ with covariance matrix $\Sigma$, and W is a distance-based sliding window comprising from 2 to 10 meters of each robot trajectory. Considering that both the UWB system and the odometry system provide estimates at a rate of $\approx$ 10Hz and there are eight available anchor-tag pairs reporting data simultaneously, each optimization problem produces in the order of $\approx$ 600 scalar UWB residuals when the robots are moving, although the number depends on the speed of the robots and the quality of the communications link.

In order to simplify the problem, we reduce it to 4-DOF by making the assumption that the roll and pitch of each robot's local odometry frame are obtained by means of the onboard IMUs, as they are able to estimate roll and pitch states very accurately even with low-cost IMUs \cite{s20020340} thanks to the integration of gyroscopes and accelerometers. Thus, only the translation components and the yaw angle are expected to exhibit significant drift. In this way, we define the state vector $\mathbf{x} = [x,y,z,\theta]^{T}$, comprised of the three relative translation components and an elemental relative rotation along the vertical axis between the odometric frames. Our formulation uses SE(3) transformation matrices that we parameterize by means of the Lie Algebra formulation from \cite{sola2018micro}. Particularly, to reduce the problem to 4-DOF, we introduce the following error function $\mathbf{e}(.)$:

\begin{equation}\label{eq:3}
\mathbf{e}(T_{1}, T_{2}) = \xi(T_{1}^{-1} T_{2})_{[1:3,6]}
\end{equation}

\noindent where $\xi(.)_{[1:3,6]}$ takes the logarithmic map of its input, and retrieves the three translational components and the last angular component.

\subsection{Radar Odometry}\label{sec:radar}

While LiDAR and vision-based methods are extensively used for localization via scan-matching or image correspondences, they are particularly sensitive to challenging environmental conditions. 4D radars, on the other hand, are robust to harsh weather and lighting conditions and can be used for localization using similar techniques. They provide range, azimuth, elevation and radial or Doppler velocity - i.e. the component of the object velocity along the sensor Line of Sight (LoS) - with each detection. This information is converted into a sparse point-cloud with per-point velocity estimates. However, the raw radar data must be first pre-processed to obtain intermediate representations that can then be used to estimate the robot positions (and the map, in the case of SLAM) in a factor graph optimization framework. With this purpose, we incorporate the existing work presented in \cite{elena2024loosely} and extend it to a multi-robot scenario. In this method, the pre-processing consists of a filter to remove false positives and dynamic objects, as well as ego-motion estimation (this is, estimating the linear velocity of the sensor platform) by fusing radial velocity and IMU data in a loosely coupled formulation that accounts for holonomic ground vehicles. 

Particularly, each Doppler velocity $v_{D}$ is expressed as the projection of the robot velocity vector $\mathbf{v}$ onto the unit vector $\mathbf{h}$ pointing towards the detection:

\begin{equation}\label{eq:v_doppler}
v_{D} = \mathbf{h} \cdot \mathbf{v} = h_{x}v_{x} + h_{y}v_{y} + h_{z}v_{z}.
\end{equation}

For a radar scan with $N$ detections, this yields a linear system solved by least squares:

\begin{equation}
\begin{bmatrix}
    v_{D,1} \\
    \vdots \\
    v_{D,N}
\end{bmatrix}
=
\begin{bmatrix}
    {h}_{1x} & {h}_{1y} & {h}_{1z} \\
    \vdots       & \vdots       & \vdots       \\
    {h}_{Nx} & {h}_{Ny} & {h}_{Nz}
\end{bmatrix}
\begin{bmatrix}
    v_x \\ v_y \\ v_z
\end{bmatrix}.
\label{eq:projection}
\end{equation}

For holonomic ground robots, roll ($\phi$) and pitch ($\theta$) data from the IMU is used to remove one DOF, leaving the following expression:

\begin{equation}
\mathbf{v} =
\begin{bmatrix}
    \cos(\theta) v_x \\
    \cos(\phi) v_y \\
    \sin(\theta) v_x + \sin(\phi) v_y
\end{bmatrix}.
\label{eq:velocity_decomposition}
\end{equation}

This formulation improves robustness by accounting for feasible velocities and reduces noise in the vertical component, where radar data is particularly noisy due to low resolution and reduced FoV. However, because multirotors can exhibit significant velocity changes in all three directions, all three linear velocity components need to be estimated independently. 

Finally, the estimated linear velocity is used to initialize registration-based scan-matching of the filtered pointclouds with Generalized ICP (GICP) \cite{segal2009generalized}. The concatenation of estimated relative transforms can be used as a standalone odometry system, or fused with additional measurements in a larger localization problem.

\subsection{Pose Graph Optimization}\label{sec:posegraph}

Pose graphs formulations constitute a widely documented approach to localization and SLAM problems \cite{grisetti2010tutorial}. A pose graph is composed of robot positions, which are represented as nodes, and observations coming from sensor measurements, which are represented as edges or factors. Edges are often relative pose measurements that serve as constraints between two consecutive nodes, in the case of odometry, or two arbitrary nodes, in the case of loop closing. Solving a pose graph consists of maximizing the posterior density $p(x\vert z)$ of states \textbf{x} given the relative measurements coming from observations \textbf{z}. If noise models are assumed to be Gaussian, solving this problem is equivalent to solving nonlinear least squares of the form:

\begin{equation}
X^{*} = \operatorname*{arg\,min}_\textbf{x}\sum_{i,j}||\mathbf{z_{ij}} - \mathbf{\hat{z}_{ij}}(\mathbf{x_{i}}, \mathbf{x_{j}})||_{\Sigma_{i}}^{2}
\end{equation}

\noindent where $\mathbf{z_{ij}}$ is a relative measurement between two nodes and $\mathbf{\hat{z}_{ij}}(\mathbf{x_{i}}, \mathbf{x_{j}})$ is the predicted measurement for a configuration of those two nodes, given that the data association is providing a consistent estimate.

Pose Graphs can be extended to multi-robot systems with a few considerations. In this paper, we adopt the concept of encounters and anchor nodes introduced in \cite{kim2010multiple}. Encounters are edges that connect the poses of two robots, and anchor nodes define the transformation between the local odometry frame of each robot and a common frame of reference. In this way, robot poses are expressed in their own local frames, and encounter constraints involve the anchor nodes to express the inter-robot measurement in a common reference frame. To reduce the problem to 4-DOF, each node is a 4D state $\mathbf{x} = [x,y,z,\theta]^{T}$ and we can make use of the relative error formulation presented in (\ref{eq:3}). This yields the following expression:

\small
\begin{multline}  
X^{*} = \operatorname*{arg\,min}_\textbf{x}\sum_{W}\Biggl[\sum_{r=1}^{R}\bigl(\phi^{r} + \sum_{i=1}^{Nr-1} ||\mathbf{e}(Z_{i,i-1}^{r},\hat{Z}_{i,i-1}^{r})||_{\Sigma_{i,i-1}}^{2}\bigr) \\ + \sum_{i,j \in C}||\mathbf{e}(Z_{ij}, \hat{Z}_{ij})||_{\Sigma_{ij}}^{2} + \phi^{1} \Biggr]
\end{multline}
\normalsize
The first term $\phi^{r}$ is a prior that anchors each one of the $R$ robots local trajectory to a starting location, which is set to the origin of its respective odometry frame. The second term groups the odometry constraints for each robot $r$ (observed relative transformation $Z^r$ and predicted relative transformation $\hat{Z^r}$) between the consecutive poses of each robot along their respective trajectories, for a total of $N_{r}$ relevant poses or keyframes. The third term refers to encounter constraints, represented by the $C$ set. In this term, $\hat{Z}_{ij}$ is the relative transformation between pose $T_{j}$ of robot $r$, and pose $T_{i}$ of robot $r'$, which is compared to the observation $Z_{ij}$. Finally, $\phi^{0}$ is a prior for one of the anchor nodes and addresses the gauge freedom in the global trajectory, which is set to align with the odometry frame of robot $1$, without loss of generality.

The optimization is performed over a spatial window W of $N_{r}$ keyframes, which are added every 0.5 meters of trajectory for each of the robots. 

In our setup, odometry constraints come either from radar, the wheel encoders in the case of the ground platform, or IMU integration regarding the UAV. As for inter-robot constraints, we make use of the relative estimation from UWB measurements introduced in \ref{sec:relative}, which can be used to infer a virtual inter-robot observation $Z_{ij}$ based on the odometry frame alignment provided by the optimization.

\section{Experiments}\label{sec:experiments} 

\begin{figure}[ht]
    \centering
    \begin{subfigure}[t]{0.48\columnwidth}
        \centering
        \includegraphics[height=4cm]{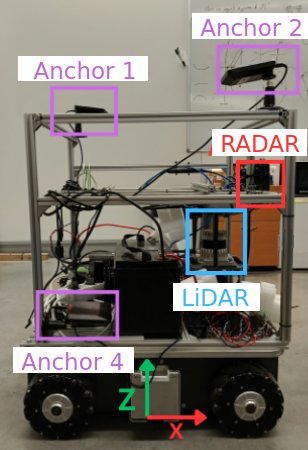}
        \caption{}
        \label{figsetup:a}
    \end{subfigure}
    \hfill
    \begin{subfigure}[t]{0.48\columnwidth}
        \centering
        \includegraphics[height=4cm]{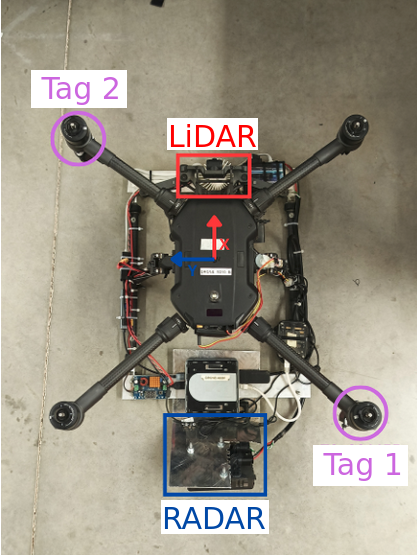}
        \caption{}
        \label{figsetup:b}
    \end{subfigure}
    \begin{subfigure}[t]{0.48\columnwidth}
        \centering
        \includegraphics[height=3cm]{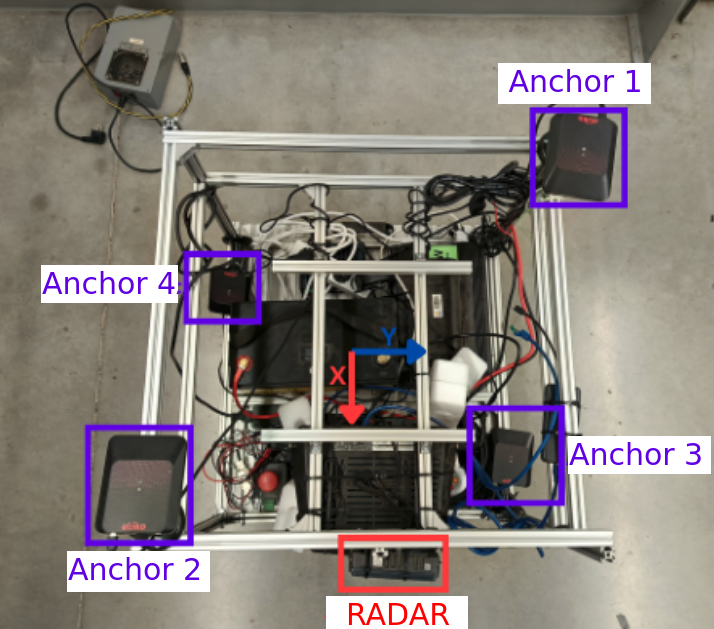}
        \caption{}
        \label{figsetup:c}
    \end{subfigure}
    \begin{subfigure}[t]{0.48\columnwidth}
        \centering
        \includegraphics[height=2.5cm]{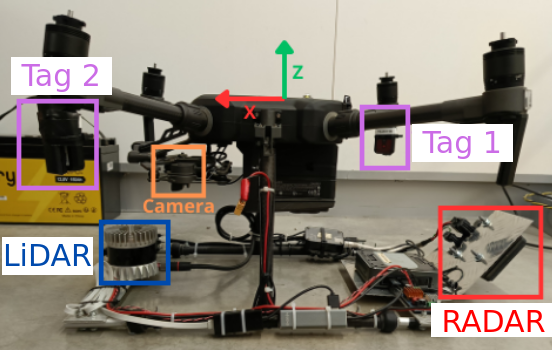}
        \caption{}
        \label{figsetup:d}
    \end{subfigure}
    \caption{(a) and (c): Top and lateral view of the UGV used in the experiments. (b) and (d): Top and lateral view of the Matrice M210 used in the experiments. UWB sensors are highlighted in purple, and the radar sensors are highlighted in red.}
    \label{fig:sensor_placement}
\end{figure}

The experiments were carried out with a holonomic UGV with a maximum payload of over 100 kilograms, and a fully autonomous DJI M210 UAV. Figure \ref{fig:sensor_placement} highlights the location of the sensors onboard both of the robots, including four Eliko UWB anchors on the UGV and two Eliko UWB tags on the UAV platform which have been replicated in simulation, as well as two ARS548RDI radars. Notice that the radar in the aerial platform is oriented backwards and tilted to point below the robot, while the radar on the UGV points forward in the same direction as the robot. Finally, both robots are equipped with OS1-32 3D LiDAR sensors, which are used to obtain the ground truth trajectories in the real setting by applying the Direct LiDAR Localization \cite{caballero2021dll} (DLL) method. 

\subsection{Ultra-Wideband characterization}\label{sec:uwb_sim}

A custom simulator plugin has been implemented to produce realistic range measurements on simulation. This plugin incorporates zero-mean Gaussian noise, affecting all measurements, and a small probability of missing samples, simulating communication issues. 

\begin{figure}[t!]
    \centering
    \includegraphics[width=0.45\textwidth]{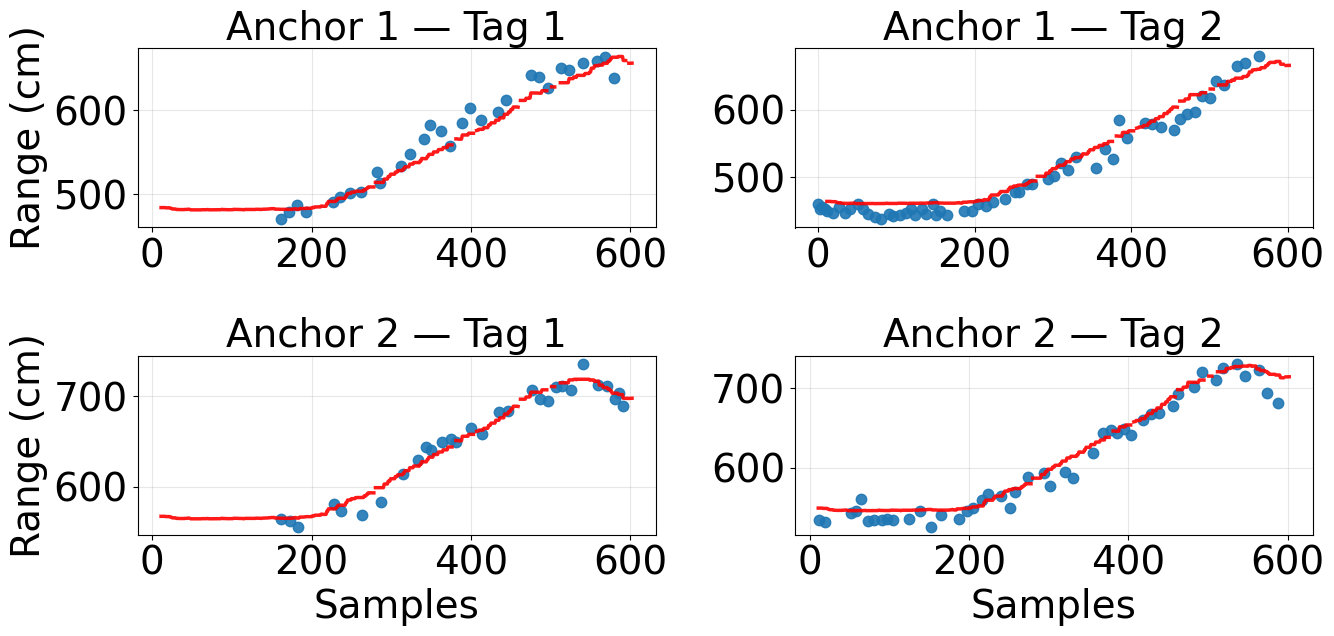}
    \caption{Measurements collected during the starting segment of the dataset trajectory for each pair \rev{(blue)}, as compared with ground truth ranges \rev{(red)}.}
    \label{fig:uwb_real_ranges}
    \vspace{-5mm}
\end{figure}

We analyzed the range measurements from two of the anchors to both tags in the real system, with the objective of adjusting the virtual measurement to model real-life behavior (see Figure \ref{fig:uwb_real_ranges}). The Mean Bias Error (MBE), the standard deviation $\sigma$ and Root Mean Squared Error (RMSE) of the real data were computed to serve as a baseline, as seen in Table \ref{tab:uwb_meas_analysis}. It was found that the average standard deviation across all pairs was under 10 centimeters in a segment when the robots were static, and slightly below 12 centimeters when they were in motion. Therefore, the standard deviation adopted for our measurement model in (\ref{eq:predicted_range}) is $\sigma = 12$ cm. Additionally, as the captured data shows a close to zero average bias across pairs under LOS conditions, we decided to model the measurements with zero bias.

Finally, the geometric distribution of tags and anchors was configured to be identical to the real-world experiments. 

\begin{table}[t!]
\centering
\vspace{2mm}
\begin{tabular}{|l|r|r|r|}
\hline
Anchor / Tag Pair & MBE [cm] & $\sigma$ [cm] & RMSE [cm] \\
\hline
Anchor 1 / Tag 1 &  8.69 & 12.55 & 15.08 \\
Anchor 1 / Tag 2 &  -9.51 & 11.63 & 14.94 \\
Anchor 2 / Tag 1 &   -0.04 & 9.00 & 8.84 \\
Anchor 2 / Tag 2 &  -6.83 & 10.93 & 12.78 \\
\hline
\end{tabular}
\caption{MBE, standard deviation, and RMSE of UWB ranges characterized on the real system.}
\label{tab:uwb_meas_analysis}
\end{table}

\subsection{Validation framework}\label{sec:validation}

\begin{figure*}[ht]
  \centering
  \begin{subfigure}{\linewidth}
    \centering
    \includegraphics[width=0.55\linewidth]{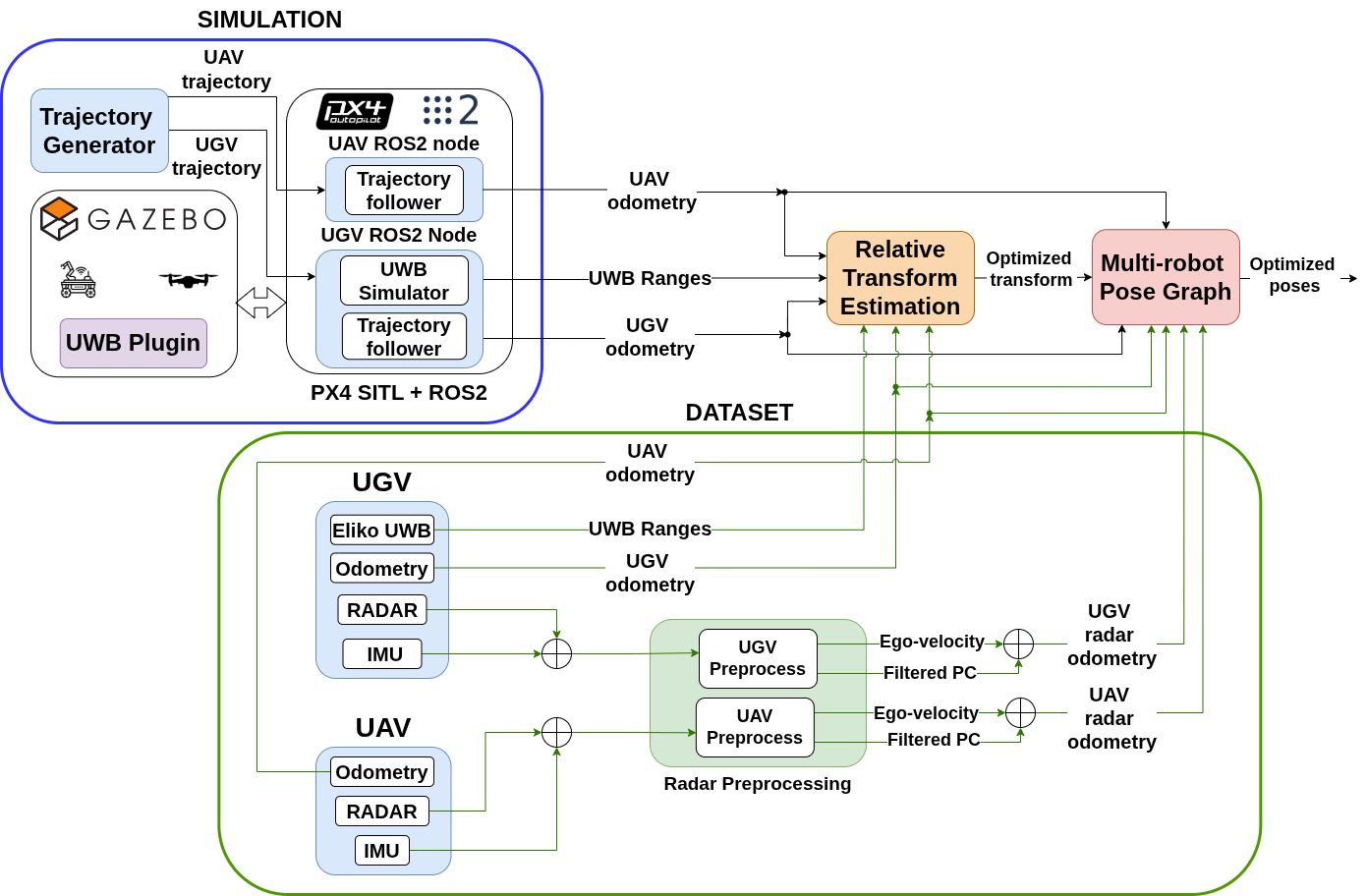}
    \caption{}
    \label{fig2:1_a}
  \end{subfigure}
  \vspace{1ex} 
  \begin{subfigure}{\linewidth}
    \centering
    \includegraphics[width=0.55\linewidth]{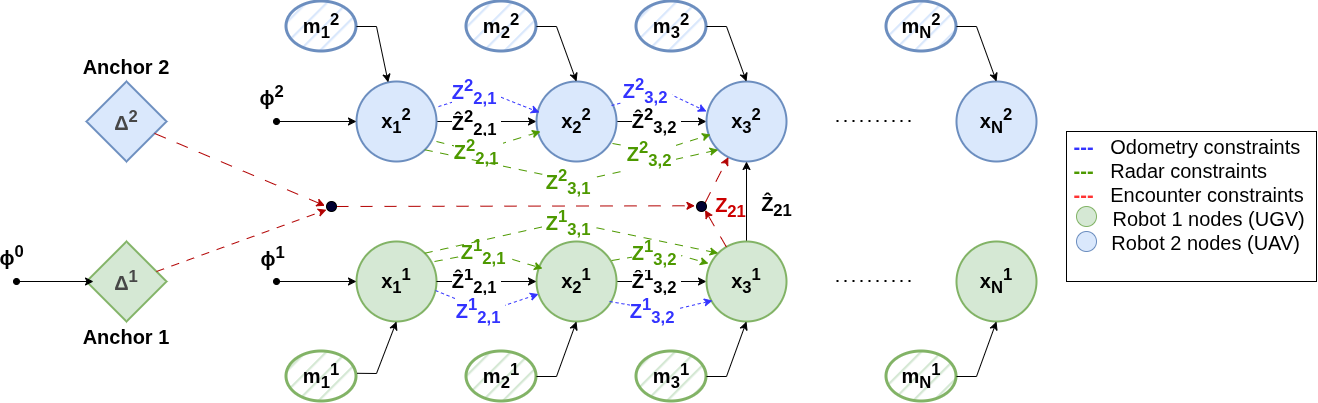}
    \caption{}
    \label{fig2:1_b}
  \end{subfigure}
  \caption{\rev{General overview of the system. (a) System architecture with the simulation setup (blue) and the real-world dataset setup (green). (b) Multi-robot Pose-Graph diagram. Each keyframe $n$ for robot $r$, $\mathbf{x}_{n}^{r}$ synchronizes the latest measurements $\mathbf{m}_{n}^r$, including odometry and radar measurements, and the virtual inter-robot observation from odometry frame alignment. }}
  \label{fig2}
\end{figure*}

The evaluation of our system has been carried out with the implementation shown in Fig. \ref{fig2} and includes two variants: the SITL simulation environment and the system developed for real-world experiments (see Fig. \ref{fig2:1_a}). Both variants share interfaces with the relative transformation estimation node and the multi-robot pose-graph node \rev{(see diagram in Fig. \ref{fig2:1_b})}, which are the main components of our system. The simulated version includes a custom simulator plugin to generate realistic virtual UWB measurements and an intermediate module that interfaces with PX4, parsing odometry into ROS 2 standard messages and commanding the robots to follow pre-generated trajectories. First, we evaluated the system in simulation without radar odometry, which is then added in the real-world setup as an additional constraint to the pose-graph framework.

\subsection{Simulation}\label{sec:simulation}

Two cooperative exploration missions were designed. In both cases, the robots start their trajectories close together ($\approx 1$ meter separation). In Mission 1, the robots describe a parametric trajectory in coordinated motion, keeping a close distance between one another. In Mission 2, the robots perform two different trajectories of varying lengths and speeds, with the aerial vehicle describing a rectangular motion and the UGV describing a circumference. This experiment aims to test the system when the robots are far apart from each other ($\approx$ 10 meters) and then get close together again. In addition to this, we modify the odometry output of both platforms, as read from the SITL, to include a 2\% drift that degrades the estimation as the robots advance.

\begin{figure}[t!]
    \centering
    \includegraphics[width=0.75\linewidth]{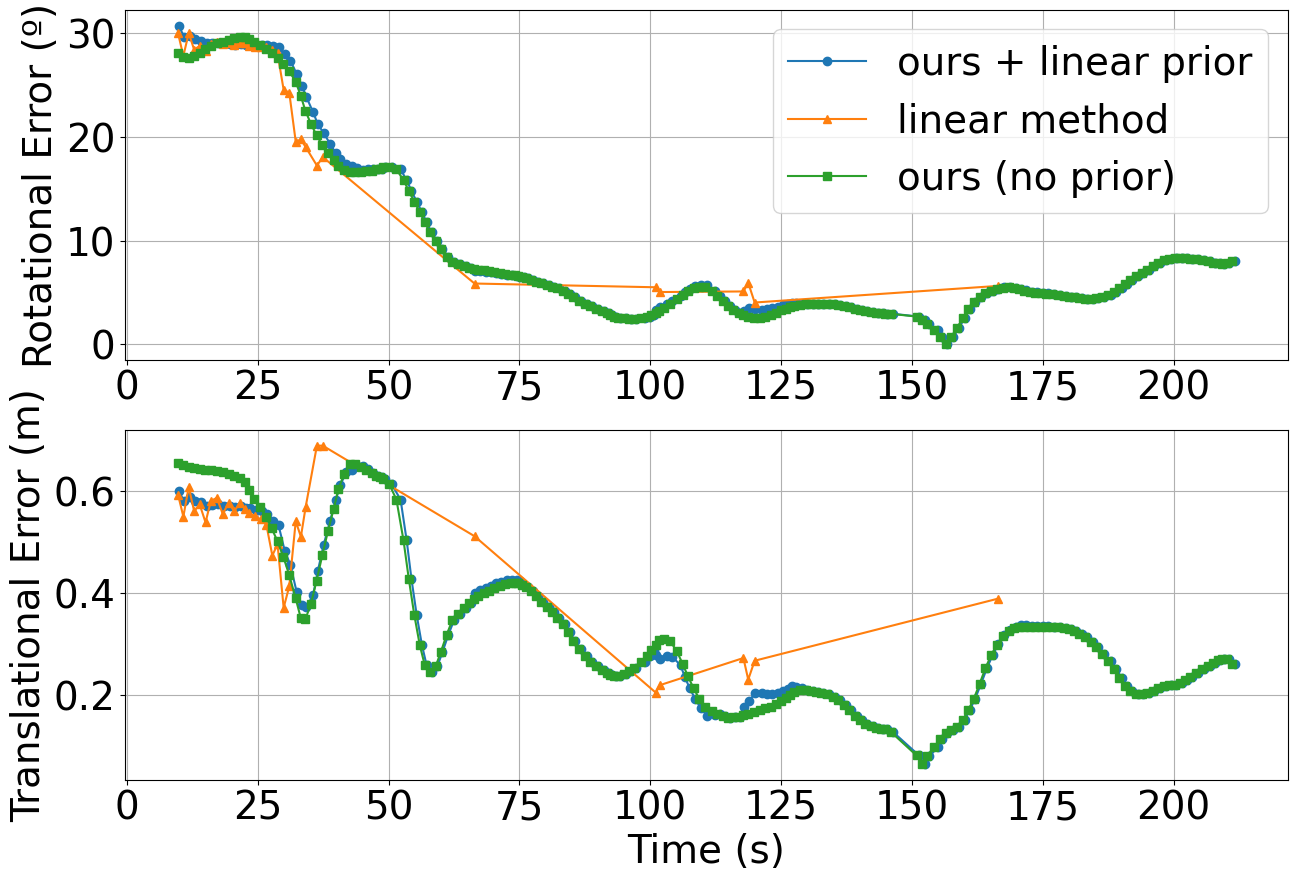}
    \caption{Comparison of the RMSE of the estimated relative transform in the three evaluated methods with $\sigma=2$ cm.}
    \label{fig:linear_method_comparison}
\end{figure}

To evaluate the 4-DOF relative transformation optimization described in Section \ref{sec:relative}, we computed the RMSE of the angular and translational components of the transformation error and compared the results with the baseline linear method from the authors of \cite{molina2019unique}, using the same standard deviation of 2 centimeters as in the original work. We selected Mission 1 as a reference, a use case where the robots stay close together. Figure \ref{fig:linear_method_comparison} compares the solution obtained by the original method, as well as our method with and without the linear solution used as prior, represented as $T_{p}$ in (\ref{eq:costfunct}). Our standalone method without an initial guess achieves a mean translational RMSE of 0.57 m and rotational RMSE of 22.9º, compared to 0.71 m and 35.5º of the linear method. Furthermore, our method provides more consistent estimates with higher convergence rates when compared to the linear method. Regarding the contribution of the linear prior to our method, there is a small benefit from the linear prior in the initial 25 seconds. However, in some cases it can undermine the performance of our method (from 115 to 125 seconds). Consequently, our standalone method is able to provide a high rate, robust estimation of the relative transform.

\begin{figure}[t!]
    \centering
    \begin{subfigure}[t]{0.48\columnwidth}
        \centering
        \includegraphics[width=\linewidth]{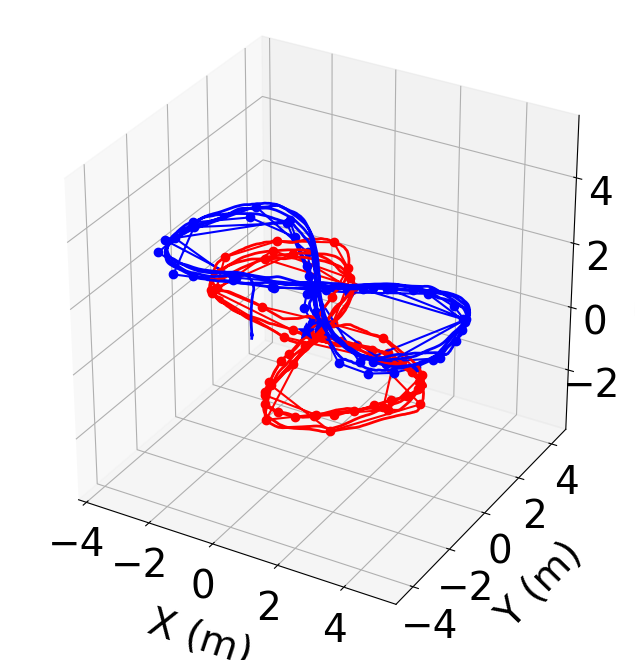}
        \caption{}
        \label{fig5:2_a_}
    \end{subfigure}
    \begin{subfigure}[t]{0.48\columnwidth}
        \centering
        \includegraphics[width=\linewidth]{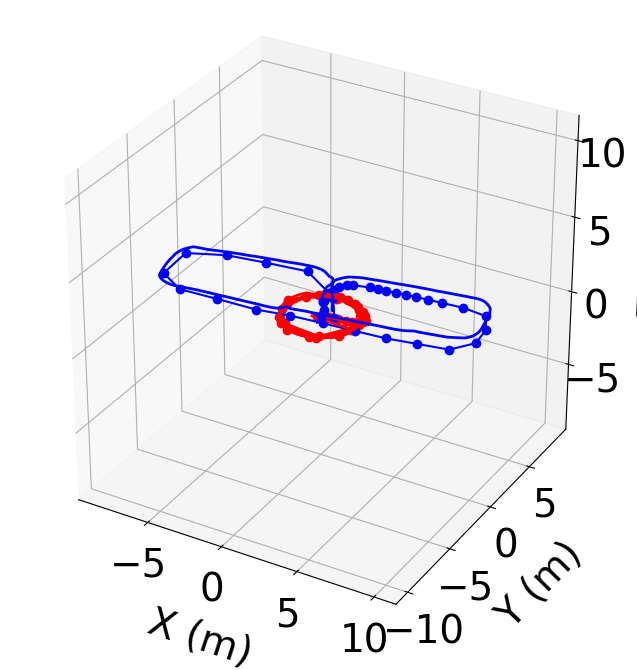}
        \caption{}
        \label{fig5:2_a__}
    \end{subfigure}
    \hfill
    \caption{Optimized poses as compared to ground truth data in Mission 1 (a) and Mission 2 (b). UGV trajectories are highlighted in blue, and UAV trajectories in red. The baseline is represented as a continuous line, and the dots represent the optimized poses.}
    \label{fig:sim_poses}
    \vspace{-6mm}
\end{figure}

Next, we evaluate the performance of our system in Missions 1 and 2 using our modeled dispersion of range measurements ($\sigma=12 $ cm). Figure \ref{fig:sim_poses} shows the reference trajectories as well as the final optimized poses for both simulated missions, with the UAV poses already transformed by the anchor nodes. In order to evaluate the algorithms, we compute the Absolute Trajectory Error (ATE) in translation, and the RMSE is used to compute the errors in the yaw angle. Table \ref{tab:ate_uncertainty_kf} shows the results of the final pose estimation for both platforms in both missions. It shows that our system is capable of estimating the global pose of both platforms even without knowledge of the initial transformation between robots. As expected, the errors on Mission 2 are higher due to DoP induced by the growing distance between the robots at the central segment of the trajectory.

\begin{table}[t!]
  \centering
  \caption{ Translational ATE in meters and RMSE of the angular error of the poses of the UAV and UGV after pose-graph optimization.}
  \label{tab:ate_uncertainty_kf}
  \begin{tabular}{|llcc|}
    \hline
    Experiment & Robot & ATE (m)& RMSE ($^\circ$) \\
    \hline
    Mission 1 & UAV & 0.98 & 25.9  \\
                            & UGV & 0.15 & 8.59 \\
    \hline
    Mission 2 & UAV & 1.51 & 27.51 \\
                             & UGV & 0.02 & 8.21\\
    \hline
    Dataset    & UAV & 1.44 & 5.74 \\
                             & UGV & 0.94 & 11.4\\
    \hline
  \end{tabular}
 \end{table}

\subsection{Dataset}\label{sec:dataset}

The dataset experiment includes a UGV trajectory that spans 27 m in the horizontal plane and includes frequent fast heading changes, and a UAV trajectory which is 118 meters long, moving in 3D space with overall smoother motions.  

\begin{figure}[t!]
    \centering
    \includegraphics[width=0.75\linewidth]{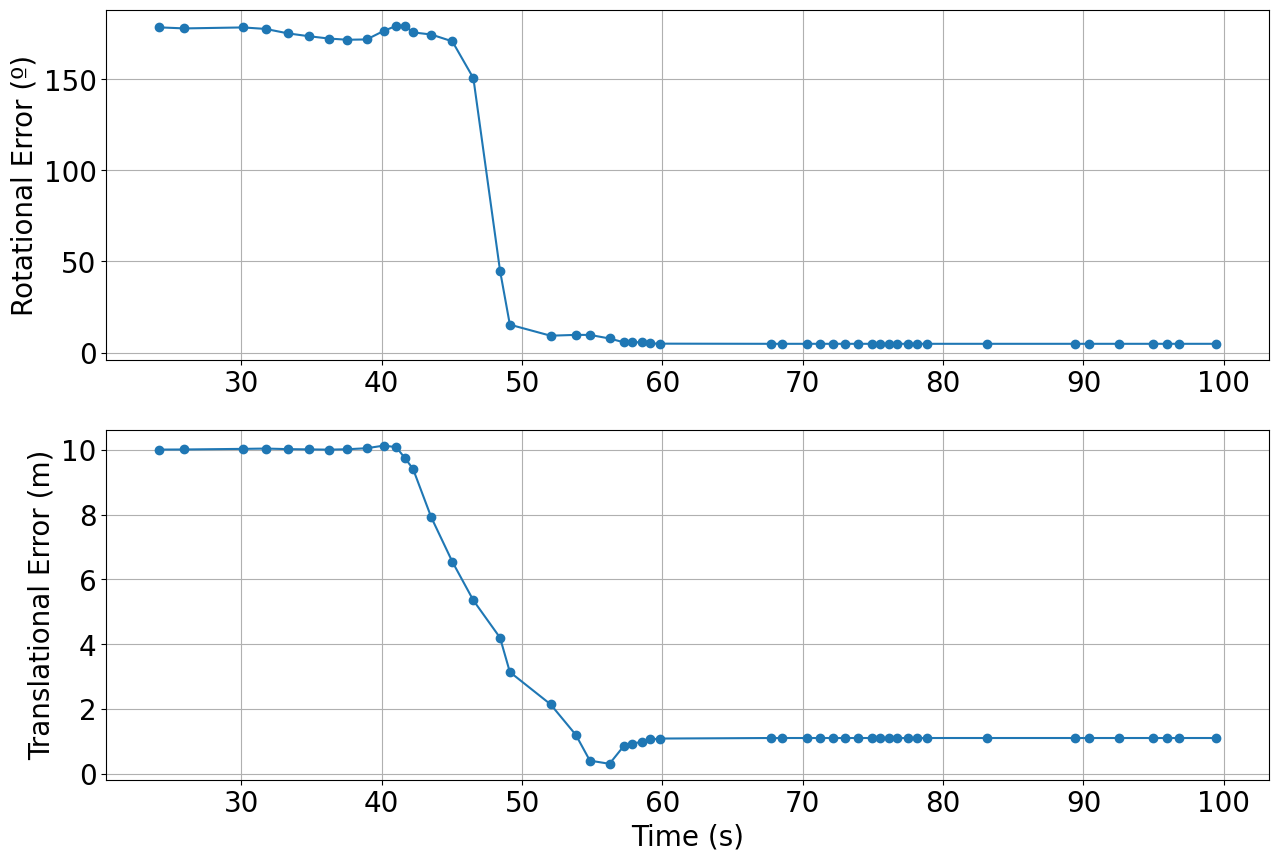}
    \caption{RMSE of the estimated relative transform with respect to ground truth from DLL, showing rotational and translational components.}
    \label{fig:dataset_relative_rmse}
    \vspace{-5mm}
\end{figure}

Figure \ref{fig:dataset_relative_rmse} indicates the rotational and translational components of the RMSE of the estimated inter-robot transformation. Approximately after 40 seconds, when both of the robots have advanced enough, the error quickly converges to $\approx$ 1.1 m in translation and $\approx$ $5^{\circ}$ in rotation. Considering measurement error, which can be on the order of tens of centimeters \rev{for the inter-robot distances in our experiments (from 0.5 to 10 meters)}, and odometry error (in this case, only wheel and IMU odometry are being fed to the optimizer, both of which tend to accumulate significant drift), the results obtained for real robots are favorable and consistent with what was obtained in simulation, which can also be confirmed by analyzing the errors of the global pose estimation (see Table \ref{tab:ate_uncertainty_kf}).

\begin{figure}[t]
  \centering
  \begin{subfigure}{\linewidth}
    \centering
    \includegraphics[width=0.75\linewidth]{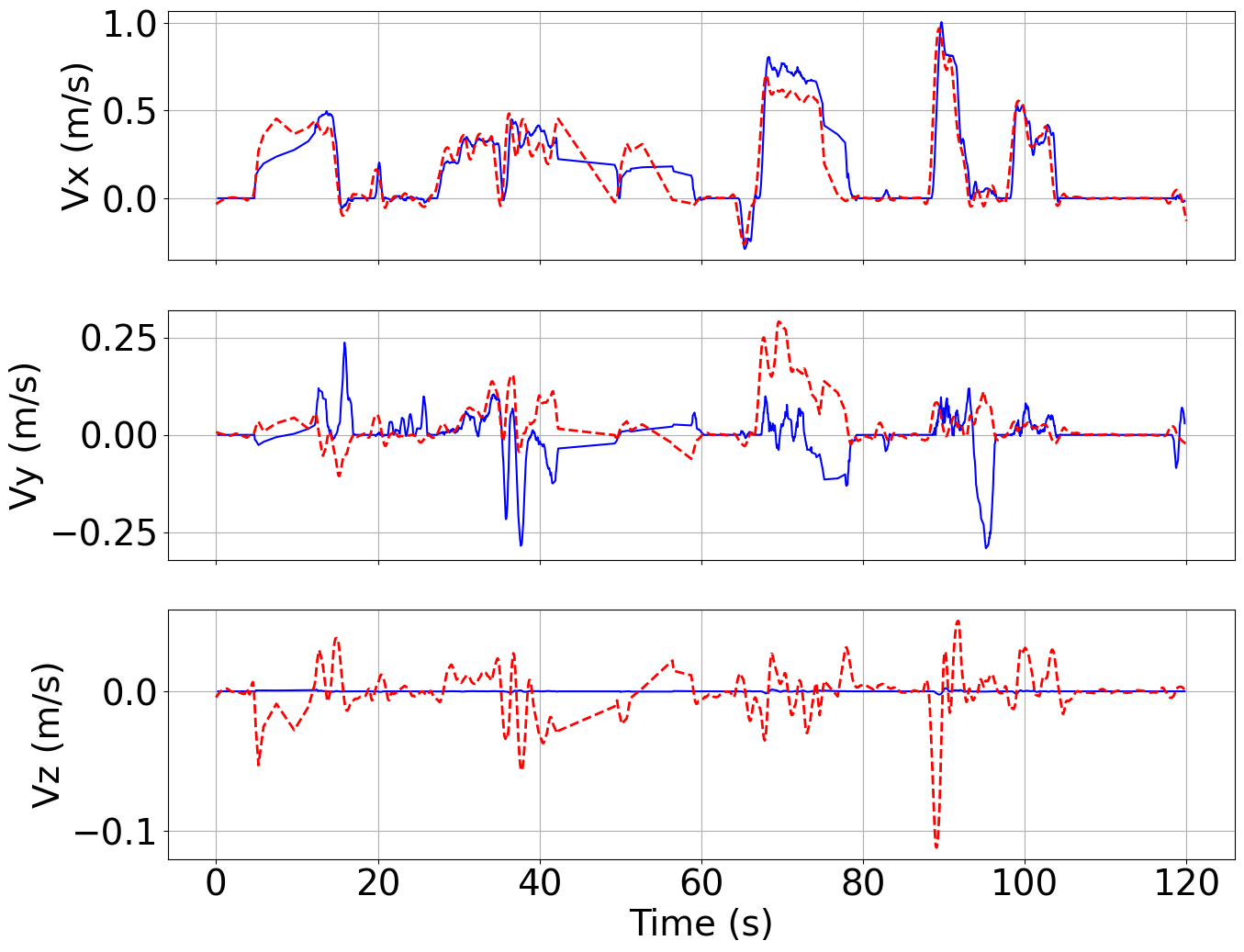}
    \caption{UGV}
    \label{fig5:9_a}
  \end{subfigure}
  \hfill
  \begin{subfigure}{\linewidth}
    \centering
    \includegraphics[width=0.75\linewidth]{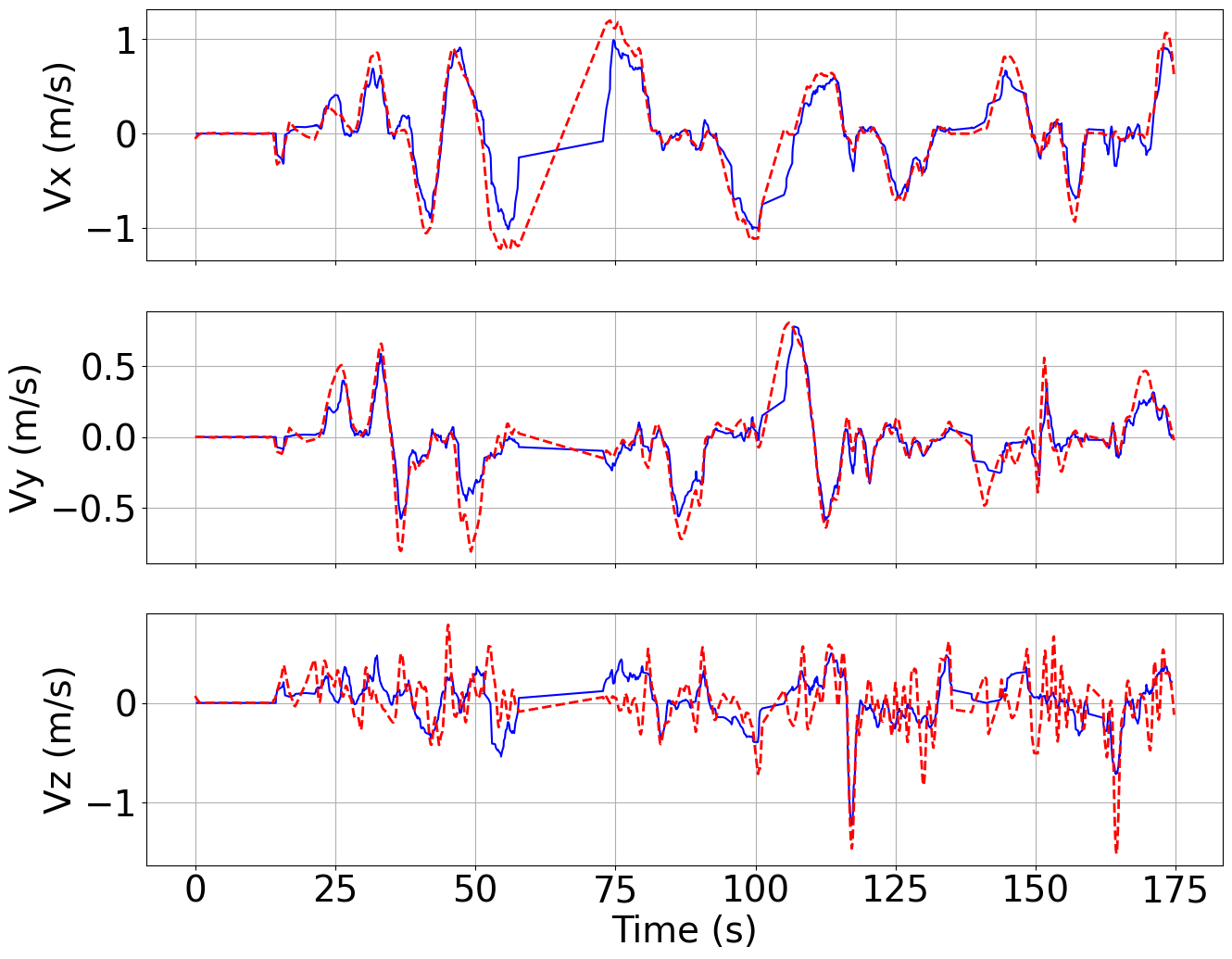}
    \caption{UAV}
    \label{fig5:9_b}
  \end{subfigure}
  \caption{Radar-estimated velocities (blue) as compared with DLL velocities (red) in the UGV (a) and the UAV (b).}
  \label{fig:radar_velocities}
\end{figure}

Figure \ref{fig:radar_velocities} represents the linear velocities estimated by the radar ego-motion algorithm, smoothed with a 10 sample moving average filter. Ground truth linear velocities are estimated by computing the derivative of the DLL pose estimation, and similarly smoothing the result. In this experiment, despite its holonomic configuration, the robot moves predominantly in the forward direction, where the radar is able to capture significant changes in velocity ($\approx$ 0.5 m/s). In the UAV case, which proves more challenging as there is individual motion in the three directions, the method is also able to provide a consistent estimate.

\begin{figure}[t!]
    \centering
    \includegraphics[width=0.75\linewidth]{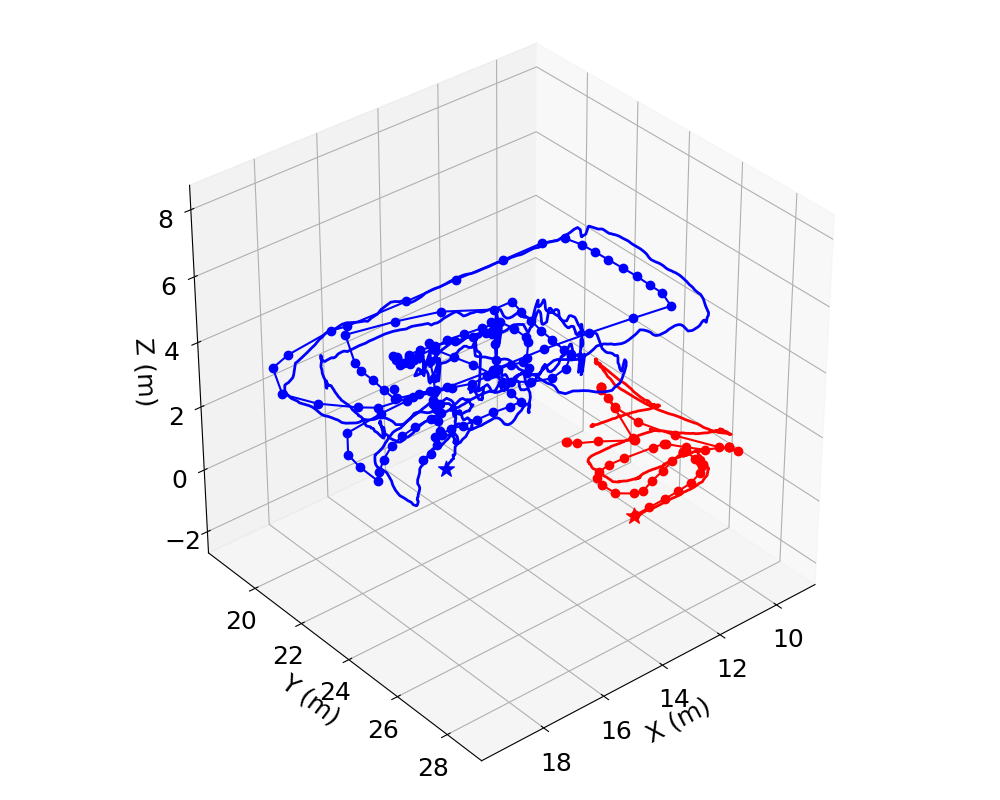}
    \caption{3D representation of the global optimized poses with anchor transformation.}
    \label{fig:dataset_3d}
    \vspace{-6mm}
\end{figure}

Figure \ref{fig:dataset_3d} shows the final optimized trajectories of the UAV and the UGV in 3D. As in the simulation scenario, the UGV trajectory is anchored to the map while the UAV trajectory is not, which means the relative localization error is manifested on the latter. Specifically, the bias in the Y direction of the UAV trajectory matches the errors found in the relative transformation ($\approx$ 1.1 m). Additionally, it was found that radar scan-matching frequently produced inconsistent translations in the vertical direction that contaminated the estimation, which was compensated by setting a higher variance in this direction to the relative transformation constraint computed by GICP. This is a well-known issue in radar-based systems due to the low vertical resolution of the sensor \cite{elena2024loosely}, and is most notably seen in the UAV trajectory.

Finally, performance metrics for all optimization experiments are provided in Table \ref{tab:performance_kf}. All experiments were performed in real time from a stream of data stored in a ROS 2 bag file, and run on a Gigabyte AORUS 5 KE laptop with a 12th Gen Intel® Core™ i7-12700H CPU and 32GB of RAM. The temporal metric is provided by Ceres and corresponds to the total time spent in the solver. The relative transformation estimation is overall less demanding due to mostly scalar residuals. In the dataset, it takes under 200 ms with a variance in the order of 100 ms. Conversely, pose-graph optimization shows more variance across experiments, which can be attributed to the variable number of 4D residuals being included in the optimization problem depending on the availability of sensor measurements. The average time per solve in the case of the dataset 
implies that in real-world operation the system would be able to run at rates above 5 Hz, as both processes execute concurrently, making it suitable for real-time localization. \rev{In our system, the computational cost of the relative transform estimation is expected to scale linearly per-pair in measurement count, at $O(MC)$ for M measurements in the window and C active pairs. As for Pose-Graph optimization, the workload will be dominated by encounters C at $O(CK^{2})$ for K active keyframes. In fully connected teams, $C = R(R-1)/2$, so the total cost will be nearly quadratic in R (total number of robots). For disjoint pairs, $C\approx R/2$ and the total cost will be near-linear in R. Quantitative validation beyond teams of two robots is left for future work.}

\begin{table}
  \centering
   \caption{Average and standard deviation of the execution time of the optimization processes for the RTE and Pose-Graph optimization problems.  The number of runs is indicated in parentheses.}
  \label{tab:performance_kf}
  \begin{tabular}{|lcc|}
    \hline
    Experiment   & RTE (s) & Pose-Graph (s) \\
    \hline
    Mission 1 & 0.31 ± 0.18 (352) & 0.77 ± 0.74 (119) \\
    Mission 2 & 0.31 ± 0.15 (168) & 0.44 ± 0.48 (94) \\
    Dataset      & 0.140 ± 0.141 (50)  & 0.118 ± 0.238 (226) \\
    \hline
    
  \end{tabular}
  \vspace{-5mm}
\end{table}

\section{Conclusions}\label{sec:conclusions}

This work showed the potential of radio-based localization methods in GPS-denied environments where LiDAR or vision may fail - e.g., in poorly lit scenes or in places with harsh weather conditions. A cooperative framework combining UWB and radar was proposed to localize a ground and an aerial platform. UWB anchor-tag measurements enabled trajectory alignment over short distances, reliably estimating translation and yaw even without an initial guess. However, performance degraded at longer ranges due to limited transceiver geometry and DoP. Alternatively, radar odometry estimates were obtained via ego-motion estimation and scan-matching, with Doppler velocity allowing consistent linear velocity estimation and aiding scan-matching initialization. These estimates were fused in a multi-robot pose-graph optimization with proprioceptive odometry, helping reduce drift and enforcing global consistency through anchor nodes and encounter constraints. Simulation and dataset results showed robust performance, with relative localization errors of approximately 1.2 \% on long, complex UAV trajectories.

However, although our SITL framework models the dispersion of range measurements from real data, a more exhaustive analysis of the bias, including behavior under Non Line Of Sight (NLOS) conditions, is left for future work. Additionally, this system does not implement explicit place recognition, which leads to growing uncertainty over time. A promising direction is extending this approach to a fully-fledged multi-robot, radio-based SLAM system, which our factor graph formulation naturally supports. However, radar place recognition remains challenging due to narrow FoV and low vertical resolution, which could be overcome by introducing more involved feature extraction methods such as learning-based embeddings \cite{peng2024transloc4d}. \rev{Finally, experiments in more challenging environments (e.g. unstructured settings or adverse weather conditions) would be necessary to further evaluate the capabilities of our system.}









\balance
\bibliographystyle{IEEEtran}
\bibliography{bibliography}

\end{document}